\documentclass[runningheads]{llncs}
\usepackage[T1]{fontenc}
\usepackage{bbding}
\usepackage{enumitem}
\usepackage{booktabs}
\usepackage{footmisc}
\usepackage{subscript}
\usepackage{setspace}
\usepackage{amsmath}
\usepackage{amssymb}
\usepackage{cite}
\usepackage{amsfonts} 
\usepackage{pifont}
\usepackage{microtype}
\usepackage{graphicx}
\usepackage{threeparttable}
\usepackage{tablefootnote}
\usepackage{color}
\usepackage{xcolor}
\usepackage{multirow}
\usepackage{algorithm}  
\usepackage{algpseudocode}
\usepackage[colorlinks,
            linkcolor=blue,
            anchorcolor=blue,
            citecolor=blue]{hyperref}
\begin{document}
%
\title{KARMA: A Multilevel Decomposition Hybrid Mamba Framework for Multivariate Long-Term Time Series Forecasting}
\titlerunning{KARMA}
%
\author{Hang Ye\inst{1,2}\textsuperscript{(\Envelope)} \and
Gaoxiang Duan\inst{1,2}\and
Haoran Zeng\inst{1,2}\and
Yangxin Zhu\inst{3}\and
Lingxue Meng\inst{1,2}\and
Xiaoying Zheng\inst{1}\textsuperscript{(\Envelope)}\and
Yongxin Zhu\inst{1}\textsuperscript{(\Envelope)}}
\authorrunning{H. Ye et al.}
%
\institute{Shanghai Advanced Research Institute, Chinese Academy of Sciences,Shanghai,China\\
\email{\{yehang, zhengxy, zhuyongxin\}@sari.ac.cn}\\
\and University of Chinese Academy of Sciences, Beijing, China\\
\and Beijing Jiaotong University, Weihai Campus, Weihai, China\\
}
\maketitle              
\begin{abstract}
Multivariate long-term and efficient time series forecasting is a key requirement for a variety of practical applications, and there are complex interleaving time dynamics in time series data that require decomposition modeling. Traditional time series decomposition methods are single and rely on fixed rules, which are insufficient for mining the potential information of the series and adapting to the dynamic characteristics of complex series. On the other hand, the Transformer-based models for time series forecasting struggle to effectively model long sequences and intricate dynamic relationships due to their high computational complexity. To overcome these limitations, we introduce KARMA, with an \textbf{A}daptive \textbf{T}ime \textbf{C}hannel \textbf{D}ecomposition module (ATCD) to dynamically extract trend and seasonal components. It further integrates a \textbf{H}ybrid \textbf{F}requency-\textbf{T}ime \textbf{D}ecomposition module (HFTD) to further decompose Series into frequency-domain and time-domain. These components are coupled with multi-scale Mamba-based KarmaBlock to efficiently process global and local information in a coordinated manner. Experiments on eight real-world datasets from diverse domains well demonstrated that KARMA significantly outperforms mainstream baseline methods in both predictive accuracy and computational efficiency. Code and full results are available at this repository: \url{https://github.com/yedadasd/KARMA}

\end{abstract}
\section{Introduction}
Time series forecasting plays a vital role in diverse real-world applications including climate prediction, financial analysis, and transportation monitoring~\cite{wu2021autoformer, Das2024state, Yan2024autoclus}. While traditional methods struggle with high-dimensional data and long-term dependencies, deep learning approaches particularly Transformer architectures~\cite{Nie2023PatchTST,liu2024itransformer, liu2022non} have demonstrated superior capability in modeling complex temporal relationships. However, their quadratic computational complexity hinders their effectiveness in long-term sequence forecasting. Recent state-space models (SSMs) like Mamba~\cite{gu2024mamba} offer a promising alternative with linear time complexity and enhanced context modeling, yet existing implementations~\cite{Wang2024IsME} inadequately address the nonlinear dynamics and multi-scale characteristics inherent in real-world time series, leaving room for further advancements.


Forecasting accuracy remains fundamentally constrained by the entangled nature of seasonal variations, stochastic noise, and dynamic trends. Although decomposition methods like STL~\cite{cleveland1990stl} and frequency analysis~\cite{cooley1965fft} enable component separation, current approaches exhibit critical rigidity: 1) Fixed decomposition rules (e.g., uniform seasonal windows) struggle with variable cycles (daily/weekly patterns in traffic vs. energy data); 2) Moreover, frequency-domain signals are frequently treated as simple supplements to the time domain~\cite{wu2021autoformer, Zhou2022fedformer}, without fully integrating time-domain properties, potentially introducing additional noise.

To address these limitations in existing time series modeling methods, this paper explores a novel direction: KARMA, which integrates multi-level dynamic decomposition with state-space model Mamba. Specifically, we propose an Adaptive Time Channel Decomposition module (ATCD) that processes the channel dimension to dynamically extract seasonal and trend components of complex time series. Additionally, we introduce a Hybrid Frequency-Temporal Decomposition module (HFTD), which further decomposes seasonal components into frequency-domain signals while incorporating temporal information. Through multi-scale modeling enabled by Mamba-based KarmaBlock, we achieve a comprehensive and efficient framework for time series forecasting. Our main contributions can be summarized as follows:
\begin{itemize}[label=\tiny$\bullet$]
    \item To the best of our knowledge, KARMA is the first framework to integrate Mamba with multi-level decomposition for multi-scaled hybrid frequency-temporal analysis, fully leveraging Mamba's potential to overcome the limitations of Transformer architectures and existing decomposition methods for efficient time series forecasting.
    \item We propose ATCD for dynamic seasonal-trend decomposition and HFTD for hybrid frequency-temporal analysis, enabling the effective utilization of latent information in time series data and improving predictive accuracy.
    \item Comprehensive experiments on eight real-world datasets across various domains demonstrate that KARMA achieves state-of-the-art performance in most benchmarks.
\end{itemize}

\section{Related Work}
\subsubsection{Deep Learning for Time Series Forecasting.}Deep learning has become pivotal in time series forecasting, with diverse architectural innovations. DeepAR~\cite{salinas2020deepar} utilizes RNN autoregressive mechanisms to model sequences while capturing seasonality and covariate dependencies. TimesNet~\cite{wu2023timesnet} transforms time series into 2D tensors and applies hierarchical convolutions. TSMixer~\cite{chen2023tsmixer} enhances multivariate modeling via channel mixing and hierarchical patch mechanisms. Transformer-based methods~\cite{Transformer}, such as PatchTST~\cite{Nie2023PatchTST}, introduce patching strategies to capture multi-scale dependencies, while Pyraformer~\cite{Liu2021Pyraformer} optimizes pyramid attention for efficient sequence modeling. Despite their advancements, the quadratic complexity of self-attention in Transformers still limits scalability. SMamba~\cite{Wang2024IsME}, leveraging linear complexity Mamba blocks, delivers strong predictive performance, yet lacks targeted optimization for the nuanced characteristics of time series data.
\vspace{-15pt}
\subsubsection{Time Series Decomposition.}Time series decomposition unveils the intricate nature of time series by extracting components such as trends and seasonality. N-BEATS \cite{oreshkin2019nbeats} and DLinear \cite{zeng2023dlinear} integrate decomposition techniques to enhance forecasting performance, but their fixed decomposition rules limit adaptability to complex sequences. Frequency-domain decomposition methods, such as Autoformer \cite{wu2021autoformer}, facilitate global modeling by extracting frequency-domain features, yet they lack joint analysis of time and frequency domains. The potential of multi-level decomposition combining time-frequency analysis remains underexplored.

\section{KARMA}
\subsection{Problem Formulation}
Without loss of generality, Given a historical sequence $X = [x_1, x_2,..., x_L] \in \mathbb{R}^{L\times D}$, where $D$ represents variates (feature channel) and $L$ represents time steps. ${x}_{t}\in\mathbb{R}^{D}$ represents containing $D$ different feature information at time stamp $t$. The forecasting task aims to predict future values $\hat{Y} = [x_{L + 1},..., x_{L + T}] \in \mathbb{R}^{T\times D}$ through a forecasting model $f_{\theta}$:
\begin{equation}
    \texttt{Time} \ \texttt{Series} \ \texttt{Forecasting}: \ \hat{\mathcal{Y}}=f_\mathrm{\theta}(\mathcal{X}),
\end{equation}
where $L\ll T$ for long-term horizons. Here $x_t\in\mathbb{R}^{D}$ denotes multivariate observations at timestamp $t$. Real-world scenarios exhibit heterogeneous temporal patterns across: (1) sampling frequencies (like hourly/daily), (2) variate cardinality $D$, and (3) feature semantics (like sensors).
\vspace{-10pt}
\begin{figure}[htbp]
\includegraphics[width=\textwidth]{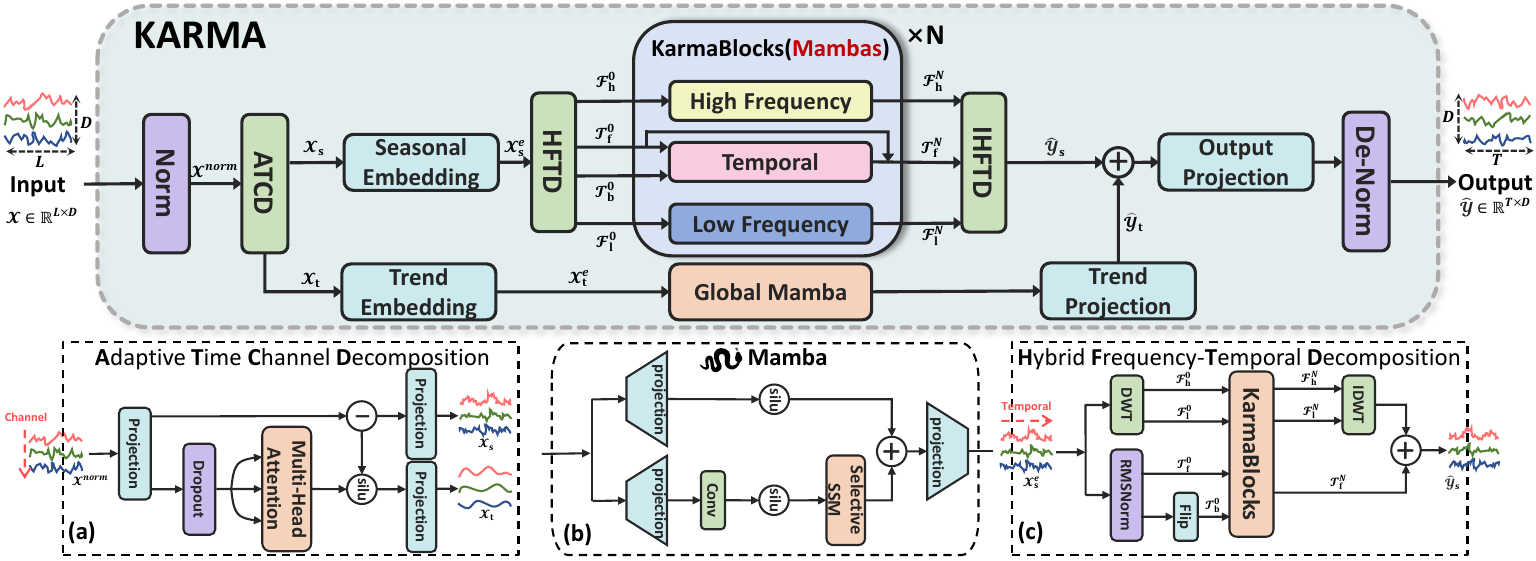}
\caption{Overall structure of KARMA. (a) ATCD module, which operates along the channel dimension. (b) Details of Mamba, the core is the selective SSM, Composed of High Frequency Mamba, Low Frequency Mamba and Temporal Mamba to form Karmablock, and Global Mamba. (c) HFTD module, integrated with KarmaBlocks, leveraging Mambas across multiple scales, which functions along the temporal dimension.} 
\label{fig2_KARMAFramework}
\vspace{-10pt}
\end{figure}
\subsection{Overview of the Framework}
The overall structure of KARMA is depicted in Fig.~\ref{fig2_KARMAFramework}. It comprises different components such as ATCD, HFTD, KarmaBlocks and preprocessing components.

Initially, the input sequence undergoes normalization and is decomposed by the ATCD into seasonal and trend components. These components are embedded via MLP-based projections to capture multi-scale temporal features. The seasonal branch employs HFTD to disentangle frequency-domain patterns and temporal dynamics, which are iteratively refined through stacked KarmaBlocks for cross-scale interaction. Concurrently, the trend component is processed by Global Mamba for long-range dependency modeling. The refined features from both pathways are then reconstructed through inverse HFTD transformation and projection operations, followed by component fusion and denormalization to generate the final prediction. This architecture explicitly models temporal patterns at multiple resolutions while maintaining linear computational complexity.

Additionally, the loss function is defined in Eq. (\ref{eq9}), which contains both time-domain loss (MSE) and an additional frequency-domain loss as inspired by \cite{Wang2024FreDFLT}. The parameter $\alpha$ is the weight to balance the time-domain and frequency-domain losses, and $\mathcal{F}$ represents the frequency transform function \cite{cooley1965fft}.
\begin{equation}
\label{eq9}
   \mathcal{L}_{\mathrm{f}} = \alpha \cdot \sum_{i=1}^{T}\left\|\mathbf{x}_{L+i}-\hat{\mathbf{x}}_{L+i}\right\|_{2}^{2} + (1 - \alpha) \cdot \sum_{i=1}^{T} \left|\mathcal{F}(\mathbf{x}_{L+i})-\mathcal{F}(\hat{\mathbf{x}}_{L+i})\right|.
\end{equation}

\subsection{Adaptive Time Channel Decomposition}\label{sec:atcd}
To address the limitations of traditional seasonal-trend decomposition in dynamically capturing trend information, we find that multi-head attention mechanism known for its ability to dynamically capture global dependencies, are highly suitable for adaptive global trend extraction from $\tilde{\mathcal{X}}^{in}$. We thus designed ATCD, as shown in Fig.~\ref{fig2_KARMAFramework}(a), the process consists of: feature mapping, attention-based trend and seasonality extraction, and output mapping. 

Given a normalized time series input $\mathcal{X}^{norm} \in \mathbb{R}^{L\times D}$, the feature dimensions are first mapped as follows:
\begin{equation}
\label{eq10}
 \tilde{\mathcal{X}}^{in} = \texttt{Dropout}(\texttt{Proj}^{di}_\theta(\mathcal{X}^{norm})),
\end{equation}

After linear projection and applying dropout, we obtain $\tilde{\mathcal{X}}^{in} \in \mathbb{R}^{L \times I}$, where $\texttt{Dropout}(*) = r \odot *, \quad r \sim \texttt{Bernoulli}(p)$. Then, adaptive global trend capture is performed on $\tilde{\mathcal{X}}^{in}$ using multi-head attention:
\begin{equation}
\label{eq11}
    \texttt{MHA}(\tilde{\mathcal{X}}^{in}) = \texttt{Concat}\left(\texttt{softmax}\left(\frac{Q_\mathrm{h} K_\mathrm{h}^\top}{\sqrt{d_\mathrm{k}}}\right)V_\mathrm{h} \;\middle|\; h \in \{1, \dots, H\} \right) W_\mathrm{O},
\end{equation}

Where $h\in\{1,\dots,H\}$ represents the number of attention heads, $d_{\mathrm{n}}$ denotes the dimensionality of the embedding vector, and the corresponding $\sqrt{d_{\mathrm{n}}}$ serves as the scaling factor. The normalized weights are obtained via the $\texttt{softmax}$ operation. $Q_h,K_h,V_h\in\mathbb{R}^{L\times d_{\mathrm{n}}}$ are computed from the input $\mathcal{X}^{in}$ using the weight matrices $W_Q,W_K,W_V\in\mathbb{R}^{L\times d_{\mathrm{n}}}$. The resulting trend are further refined through the activation function, yielding the raw trend components. Subsequently, the raw seasonal components can be derived:
\begin{equation}
\label{eq12}
    \tilde{\mathcal{X}}_{\mathrm{t}}  = \texttt{SiLU}(\texttt{MHA}(\tilde{\mathcal{X}}^{in})),\ \tilde{\mathcal{X}}_{\mathrm{s}}  = \tilde{\mathcal{X}}^{in} - \tilde{\mathcal{X}}_{\mathrm{t}},
\end{equation}

The final seasonal and trend components are obtained by linear projection:
\begin{equation}
\label{eq13}
    \mathcal{X}_{\mathrm{t}} = \texttt{Proj}^{do\_t}_\theta(\tilde{\mathcal{X}}_{\mathrm{t}}), \
    \mathcal{X}_{\mathrm{s}} = \texttt{Proj}^{do\_s}_\theta(\tilde{\mathcal{X}}_{\mathrm{s}}). 
\end{equation}

\subsection{Hybrid Frequency-Temporal Decomposition with KarmaBlocks}\label{sec:km_hftd}
To further extract seasonal component information, we developed HFTD combined with karmablocks. As shown in Fig.~\ref{fig2_KARMAFramework}(c), HFTD not only overcomes the quadratic computational complexity of Transformers but also fully leverages the temporal and frequency characteristics of time series. The process involves: decomposing the embedded seasonal components into hybrid time-frequency representations. Simultaneously, multi-scale ensemble Mambas are stacked within KarmaBlocks to process the decomposed time-frequency signals. Ultimately, the signals are inversely transformed and output.

For frequency-domain characteristic extraction, we utilize Discrete Wavelet Transform~\cite{daubechies1990wavelet} to decompose the signal into high-frequency and low-frequency components, enabling the capture of features across different frequency bands. Given the input \( \mathcal{X}^e_{\mathrm{s}} \in \mathbb{R}^{D \times E_s} \), the general form is expressed as:
\begin{equation}
\label{eq14}
\mathcal{X}_{\mathrm{s}}^{e}[d,n]=\sum_{k}\mathcal{F}_{\mathrm{l}}[j_0,k]\cdot\phi_{j,k}[n]+\sum_{j=j_0}^J\sum_{k}\mathcal{F}_{\mathrm{h}}[j,k]\cdot\psi_{j,k}[n],
\end{equation}
where $d\in\{1,\cdots,D\}$ represents the feature channel index, and $n\in\{1,\cdots,E_s\}$ denotes the time step index, and the sequence is decomposed by wavelet on a dimension-by-dimension basis. With $k\in\{1,\cdots,M\}$ indicating the shift parameter after discrete wavelet transform, depending on $E_s$. The scale $j_0 = 1,\ j\in\{1,\cdots,J\}$ corresponds to different levels of decomposition. The coefficients $\mathcal{F}_{\mathrm{l}}[j_0,k],\ \mathcal{F}_{\mathrm{h}}[j,k]\ $represent the low-frequency and high-frequency components, respectively. The basis functions for capturing the low-frequency scale and the high-frequency wavelet components are defined as follows:
\begin{equation}
\label{eq15}
    \phi_{j,k}[n] = 2^{-\frac{j}{2}}\phi\left(2^{-j}n-k\right),\
     \psi_{j,k}[n] = 2^{-\frac{i}{2}}\psi\left(2^{-j}n-k\right), 
\end{equation}
varies according to the selection of wavelet types. In this paper, we set $J = 1$ and mean only extract one high-frequency and one low-frequency component:
\begin{equation}
\label{eq16}
    \mathcal{F}_{\mathrm{h}}[k]   = \sum_{n=1}^{E_s}\mathcal{X}_{\mathrm{s}}^e[d,n]\cdot \psi_{j,k}[n], \
   \mathcal{F}_{\mathrm{l}}[k]   = \sum_{n=1}^{E_s}\mathcal{X}_{\mathrm{s}}^e[d,n]\cdot \phi_{j,k}[n], 
\end{equation}

Finally, two frequency components are obtained, i.e. $\mathcal{X}_{\mathrm{s}}^{e} \to \{\mathcal{F}_{\mathrm{h}} \in \mathbb{R}^{D \times M}, \mathcal{F}_{\mathrm{l}} \in \mathbb{R}^{D \times M}\}$. Furthermore, temporal components are incorporated, detailed as follows:
\begin{equation}
\label{eq17}
    \begin{split}
    \mathcal{T}_{\mathrm{f}} = \texttt{RMSNorm}(\mathcal{X}_{\mathrm{s}}^{e}), \ \mathcal{T}_{\mathrm{b}} = \texttt{Flip}_D(\mathcal{T}_{\mathrm{f}}),
    \end{split}
\end{equation}
Where $\mathcal{T}_{\mathrm{f}}$ can be considered as a special form of residual, retaining the original temporal information through RMSNorm. Meanwhile, inspired by~\cite{Wang2024IsME}, $\mathcal{T}_{\mathrm{b}}$ reverses the feature channel dimension of the original sequence as a form of data augmentation, enabling the learning of more diverse feature relationships.

Upon obtaining the hybrid time-frequency series through decomposition, we proceed to process them using KarmaBlocks. Within each KarmaBlock:
\begin{equation}
\label{eq18}
    \mathcal{F}_{\mathrm{h}}^{n}  = \texttt{Mamba}^{HF}_\theta(\mathcal{F}_{\mathrm{h}}^{n-1}), \ \mathcal{F}_{\mathrm{l}}^{n} =  \texttt{Mamba}^{LF}_\theta(\mathcal{F}_{\mathrm{l}}^{n-1}),
\end{equation}
\begin{equation} 
\label{eq19}
  \resizebox{0.92\hsize}{!}{$\begin{aligned}
  \mathcal{T}_{\mathrm{f}}^{n}  = \texttt{Mamba}^{T}_\theta(\texttt{RMSNorm}(\mathcal{T}_{\mathrm{f}}^{n-1})) + \texttt{Mamba}^{T}_\theta(\mathcal{T}_{\mathrm{b}}^{n-1}) + \mathcal{T}_{\mathrm{f}}^{n-1}, \ \mathcal{T}_{\mathrm{b}}^{n}\ = \texttt{Flip}_D(\mathcal{T}_{\mathrm{f}}^{n}),
  \end{aligned}$}
\end{equation}

Among them, $\texttt{Mamba}^{HF}\theta$, $\texttt{Mamba}^{LF}\theta$, and $\texttt{Mamba}^{T}_\theta$ denote the multi-scale Mambas tailored for processing distinct components. The cornerstone of the Mamba model is the discrete state-space equation:
\begin{equation}
\label{eq20}
    \mathcal{H}_{t} = \overline{\mathcal{A}} \mathcal{H}_{\mathrm{t-1}} + \overline{\mathcal{B}} \mathcal{X}_\mathrm{t}
    , \
    \mathcal{Y}_\mathrm{t} = \mathcal{C} \mathcal{H}_\mathrm{t} + \mathcal{D} \mathcal{X}_\mathrm{t},
\end{equation}  
\begin{equation}
\label{eq21}
    \overline{\mathcal{A}} = \texttt{exp}(\Delta \mathcal{A})
    \qquad
    \overline{\mathcal{B}} = (\Delta \mathcal{A})^{-1} (\texttt{exp}(\Delta\mathcal{A}) - \mathbf{I}) \cdot \Delta \mathcal{B},
\end{equation}

More generally, given an input $\mathcal{X} \in \mathbb{R}^{D \times L}$, where $\overline{\mathcal{A}} \in \mathbb{R}^{D \times L \times N}$ denotes a specially initialized state transition matrix, which is central to long-sequence modeling, $\overline{\mathcal{B}} \in \mathbb{R}^{D \times L \times N}$ represents the input matrix for processing inputs. These are derived by discretizing $\mathcal{A} \in \mathbb{R}^{L \times N}$ and $\mathcal{B} \in \mathbb{R}^{D \times N}$ using the control parameter $\Delta \in \mathbb{R}^{D \times L}$. Additionally, $\mathcal{C} \in \mathbb{R}^{D \times N}$ signifies the output matrix that governs the output content, and $\mathcal{D}\in \mathbb{R}^{D \times N}$ functions akin to a skip connection mechanism. Building upon this, Mamba adapts to data variations by employing linear mappings for these matrices and parameters (selection mechanism).

Ultimately, the raw series is obtained by combining the hybrid information modeled through KarmaBlocks with Eq. (\ref{eq14}), which is the Inversed HFTD:
\begin{equation}
\label{eq22}
    \hat{\mathcal{Y}}_{\mathrm{s}}=\sum_{k}\mathcal{F}^{N}_{\mathrm{l}}[k]\cdot\phi_{j,k}[n]+\sum_{k}\mathcal{F}^{N}_{\mathrm{h}}[k]\cdot\psi_{j,k}[n]+\mathcal{T}_{\mathrm{f}}^{N}.
\end{equation}

\section{Experiments}
\subsection{Experimental Details}
\subsubsection{Datasets and Baselines.}
The datasets we utilized are detailed in Table \ref{tab:dataset}, encompassing eight real-world datasets across four domains, including: electricity consumption, daily exchange rates, road occupancy rates, meteorological indicator sequences, and the oil temperature and power load of electrical transformers.
\begin{table}[htbp]
\vspace{-10pt}
\centering
\caption{Datasets Details. $Channels$ denotes the number of features. $Time\ steps$ is the number of samples. $Frequency$ is the sampling frequency of each sample.}
\label{tab:dataset}
\begin{small}
\setlength{\tabcolsep}{6pt}
\renewcommand{\arraystretch}{0.1}
\resizebox{0.8\columnwidth}{!}{

\begin{tabular}{c|c|c|c|c}
    \toprule
    \textbf{Dataset} & \textbf{Channels} & \textbf{Temporal Steps} & \textbf{Frequency}& \textbf{Domain} \\
    \toprule
    ECL~\cite{zhou2021informer} & 321  & 26304 & Hourly & Electricity \\
    \midrule 
    Exchange~\cite{Lai2017ModelingLA} & 8  & 7588 & Daily & Economy \\
    \midrule
    Traffic~\cite{wu2021autoformer} & 862  & 17545 & Hourly & Transportation \\
    \midrule
    Weather~\cite{wu2021autoformer} & 21  & 52697 & 10min & Weather\\
    \midrule
    ETTh1, ETTh2 & 7 & 17420 & Hourly & Electricity\\
    \midrule
    ETTm1, ETTm2~\cite{wu2021autoformer} & 7 & 69680 & 15min & Electricity\\
    \bottomrule
\end{tabular}

}
\end{small}
\vspace{-10pt}
\end{table}

Besides, we carefully selected seven representative models based on different deep learning model divisions as our baseline models. Among them, Mamba-based model: SMamba\cite{Wang2024IsME}; Transformer-based models: iTransformer\cite{liu2024itransformer}, PatchTST\cite{Nie2023PatchTST}, FEDformer\cite{Zhou2022fedformer}, Autoformer\cite{wu2021autoformer}; MLP-based model: DLinear\cite{zeng2023dlinear}; CNN-based model: TimesNet\cite{wu2023timesnet}. Among them, DLinear, FEDformer and Autoformer adopt series decomposition methods.

\subsubsection{Implementation details.}
All experiments are implemented using PyTorch 2.3.0 and run on two NVIDIA A30 24GB GPUs. The model is trained using the Adam optimizer with the loss function introduced in Eq. (\ref{eq9}). $\alpha$ is set to 0.2, which is optimal in comprehensive situations. For different datasets and different hyperparameter implementations, the general settings include setting the learning rate to $10^{-3}$, adopting a learning rate decay strategy, and setting the batch size to 32. Training is done for 10 epochs with early stopping strategy.

\subsection{Main Results}
\setcounter{footnote}{0}
As shown in Table \ref{tab:main_results}, the averaged prediction results across eight real-world datasets demonstrate that KARMA outperforms seven representative baseline methods on most datasets.
\vspace{-15pt}
\begin{table}[htbp]
  \caption{Averaged forecasting results with prediction lengths $T\in\{96, 192, 336, 720\}$ and fixed lookback length $L=96$. Full results are can be seen at repository.}
  \vspace{-5pt}
  \renewcommand{\arraystretch}{0.85} 
  \centering
  \resizebox{0.8\columnwidth}{!}{
  \begin{threeparttable}
  \begin{small}
  \renewcommand{\multirowsetup}{\centering}
  \setlength{\tabcolsep}{1.45pt}
  \label{tab:main_results}
    \begin{tabular}{c|cc|cc|cc|cc|cc|cc|cc|cc}
    \toprule
    \multirow{2}{*}{Models} & 
    \multicolumn{2}{c}{\rotatebox{0}{\scalebox{0.8}{\textbf{KARMA}}}} &
    \multicolumn{2}{c}{\rotatebox{0}{\scalebox{0.8}{SMamba}}} &
    \multicolumn{2}{c}{\rotatebox{0}{\scalebox{0.8}{iTransformer}}} &
    \multicolumn{2}{c}{\rotatebox{0}{\scalebox{0.8}{PatchTST}}} &
    \multicolumn{2}{c}{\rotatebox{0}{\scalebox{0.8}{TimesNet}}} &
    \multicolumn{2}{c}{\rotatebox{0}{\scalebox{0.8}{DLinear}}} &
    \multicolumn{2}{c}{\rotatebox{0}{\scalebox{0.8}{FEDformer}}} &
    \multicolumn{2}{c}{\rotatebox{0}{\scalebox{0.8}{Autoformer}}} \\
    &
    \multicolumn{2}{c}{\scalebox{0.8}{\textbf{(Ours)}}} &
    \multicolumn{2}{c}{\scalebox{0.8}{(2024)}} & 
    \multicolumn{2}{c}{\scalebox{0.8}{(2024)}} & 
    \multicolumn{2}{c}{\scalebox{0.8}{(2023)}} & 
    \multicolumn{2}{c}{\scalebox{0.8}{(2023)}} & 
    \multicolumn{2}{c}{\scalebox{0.8}{(2023)}} & 
    \multicolumn{2}{c}{\scalebox{0.8}{(2022)}} &
    \multicolumn{2}{c}{\scalebox{0.8}{(2021)}} \\ 
    \cmidrule(lr){2-3} \cmidrule(lr){4-5} \cmidrule(lr){6-7} \cmidrule(lr){8-9}\cmidrule(lr){10-11} \cmidrule(lr){12-13} \cmidrule(lr){14-15} \cmidrule(lr){16-17} 
    {Metric}  & \scalebox{0.78}{MSE} & \scalebox{0.78}{MAE}  & \scalebox{0.78}{MSE} & \scalebox{0.78}{MAE}  & \scalebox{0.78}{MSE} & \scalebox{0.78}{MAE}  & \scalebox{0.78}{MSE} & \scalebox{0.78}{MAE}  & \scalebox{0.78}{MSE} & \scalebox{0.78}{MAE}  & \scalebox{0.78}{MSE} & \scalebox{0.78}{MAE} & \scalebox{0.78}{MSE} & \scalebox{0.78}{MAE} & \scalebox{0.78}{MSE} & \scalebox{0.78}{MAE} \\
    \toprule
    \scalebox{0.95}{ECL} & \scalebox{0.78}{\textbf{0.168}} & \scalebox{0.78}{\textbf{0.261}} & \scalebox{0.78}{\underline{0.170}} & \scalebox{0.78}{\underline{0.267}} & \scalebox{0.78}{0.175} & \scalebox{0.78}{\underline{0.267}} & \scalebox{0.78}{0.205} & \scalebox{0.78}{0.290} & \scalebox{0.78}{0.192} & \scalebox{0.78}{0.295} & \scalebox{0.78}{0.212} & \scalebox{0.78}{0.300} & \scalebox{0.78}{0.214} &
    \scalebox{0.78}{0.327} & \scalebox{0.78}{0.245} & \scalebox{0.78}{0.334} \\ 
    
    \midrule
    \scalebox{0.95}{Exchange} & \scalebox{0.78}{\underline{0.360}} & \scalebox{0.78}{\textbf{0.404}} & \scalebox{0.78}{0.364} & \scalebox{0.78}{\underline{0.407}} & \scalebox{0.78}{0.364} & \scalebox{0.78}{\underline{0.407}} & \scalebox{0.78}{0.367} & \scalebox{0.78}{\textbf{0.404}} &
    \scalebox{0.78}{0.416} & \scalebox{0.78}{0.443} & \scalebox{0.78}{\textbf{0.354}} & \scalebox{0.78}{0.414} & \scalebox{0.78}{0.519} &
    \scalebox{0.78}{0.429} & \scalebox{0.78}{0.496} & \scalebox{0.78}{0.494} \\ 
    
    \midrule 
    \scalebox{0.95}{Traffic} & \scalebox{0.78}{0.453} & \scalebox{0.78}{0.284} & \scalebox{0.78}{\textbf{0.414}} & \scalebox{0.78}{\textbf{0.275}} & \scalebox{0.78}{\underline{0.422}} & \scalebox{0.78}{\underline{0.282}} & \scalebox{0.78}{0.481} & \scalebox{0.78}{0.304} & 
     \scalebox{0.78}{0.620} & \scalebox{0.78}{0.336} & \scalebox{0.78}{0.625} & \scalebox{0.78}{0.383} & \scalebox{0.78}{0.610} &
     \scalebox{0.78}{0.376} & \scalebox{0.78}{0.625} & \scalebox{0.78}{0.392} \\ 

    \midrule 
    \scalebox{0.95}{Weather} & \scalebox{0.78}{\textbf{0.250}} & \scalebox{0.78}{\textbf{0.277}} & \scalebox{0.78}{\underline{0.252}} & \scalebox{0.78}{\textbf{0.277}} & \scalebox{0.78}{0.259} & \scalebox{0.78}{\underline{0.280}} & \scalebox{0.78}{0.259} & \scalebox{0.78}{0.281} &
     \scalebox{0.78}{0.259} & \scalebox{0.78}{0.287} & \scalebox{0.78}{0.265} & \scalebox{0.78}{0.317} & \scalebox{0.78}{0.309} &
     \scalebox{0.78}{0.360} & \scalebox{0.78}{0.355} & \scalebox{0.78}{0.392} \\
    
    \midrule
    \scalebox{0.95}{ETT\tablefootnote{ Due to space limitations, we present the \textbf{averaged} results of the four ETT datasets. However, we achieved SOTA performance across all ETT datasets in actuality.}} & \scalebox{0.78}{\textbf{0.367}} & \scalebox{0.78}{\textbf{0.387}} & \scalebox{0.78}{0.382} & \scalebox{0.78}{0.400} & \scalebox{0.78}{0.383} & \scalebox{0.78}{0.400} & \scalebox{0.78}{\underline{0.381}} & \scalebox{0.78}{\underline{0.397}} &
     \scalebox{0.78}{0.391} & \scalebox{0.78}{0.404} & \scalebox{0.78}{0.442} & \scalebox{0.78}{0.444} & \scalebox{0.78}{0.408} &
     \scalebox{0.78}{0.428} & \scalebox{0.78}{0.452} & \scalebox{0.78}{0.455} \\ 
    
    \bottomrule
  \end{tabular}
    \end{small}
  \end{threeparttable}
 }
\end{table}
\vspace{-10pt}
These datasets' distinct trend and seasonal features validate the effectiveness of the multilevel decomposition framework in capturing relevant information from different components. In contrast, Transformer-based methods perform poorly, likely due to overfitting caused by directly processing the raw sequences and the excessive number of FFN layers in their design. Additionally, KARMA achieves the best performance on the highly periodic ECL dataset, primarily owing to the contribution of the HFTD module, as confirmed in the ablation experiments (Section \ref{sec:ablation}). However, in high-dimensional time series (such as the Traffic dataset), KARMA shows a decline in performance, which we speculate may be due to the extra noise introduced by ATCD in the high-dimensional decomposition.
\begin{figure}[htbp]
\vspace{-10pt}
\centering
\includegraphics[width=0.8\columnwidth]{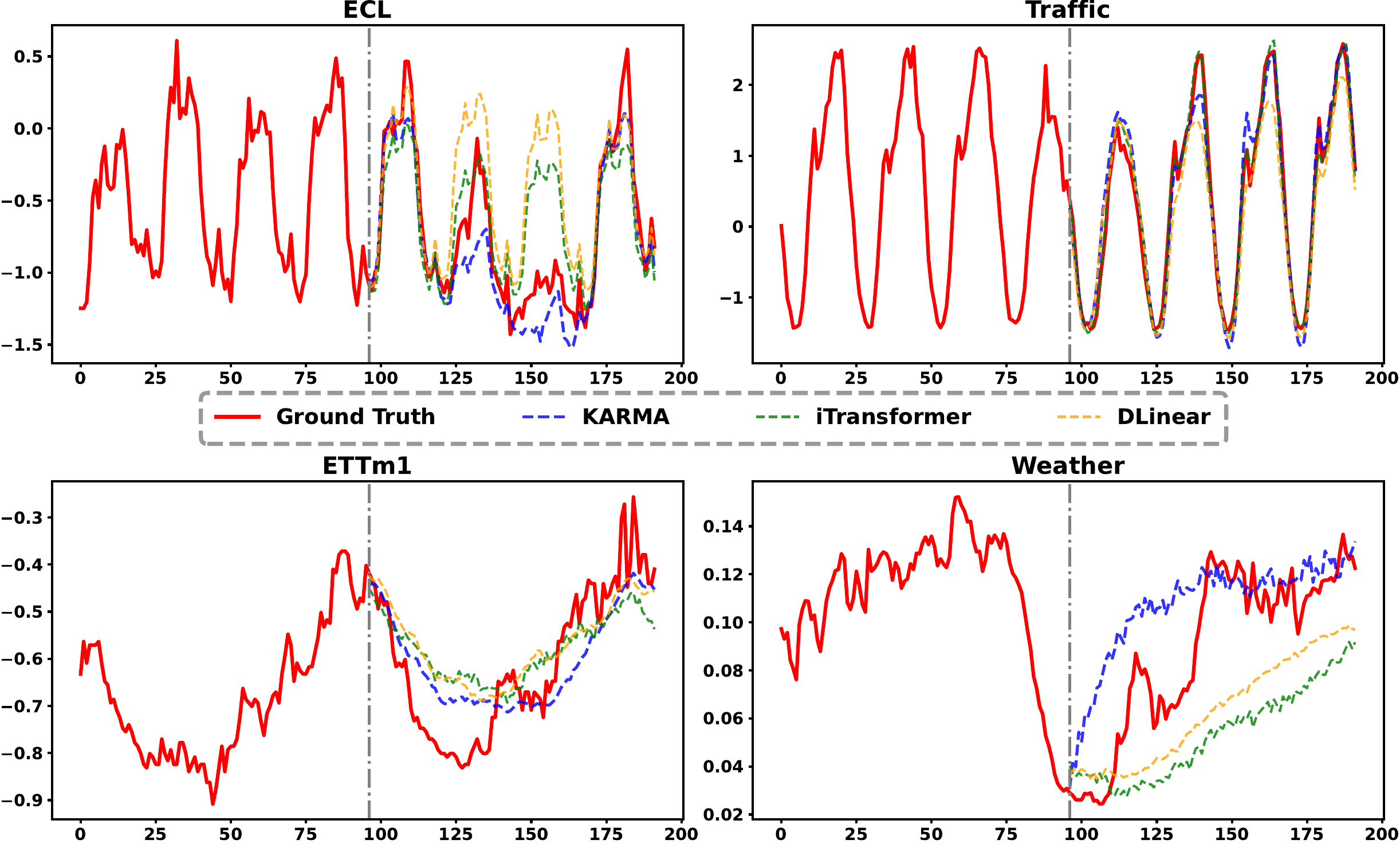}
\caption{Visualization results of input-96-predict-96 results on ECL, Traffic, ETTm1, and Weather datasets.} 
\label{fig3_results}
\vspace{-10pt}
\end{figure}

Fig.~\ref{fig3_results} illustrates the forecasting results on four representative datasets. The visual comparison clearly demonstrates that KARMA captures future trend changes more accurately than other methods, showcasing its superior performance. By combining the results from Table~\ref{tab:main_results} and Fig.~\ref{fig3_results}, it can be concluded that KARMA achieves competitive performance in long-term time series forecasting.

\subsection{Ablation Studies and Analysis}\label{sec:ablation}
\subsubsection{Ablation study.}
To further validate the effectiveness of our method, we conducted an ablation study on the multi-level decomposition components, ATCD and HFTD, as well as the KarmaBlock. Specifically, we evaluated the contributions of these modules through component replacement (KARMA$_{replace}$) or removal (KARMA$_{w/o}$). Experimental results demonstrate that KARMA generally achieves optimal performance when incorporating the multi-level decomposition components. 
\begin{table}[htbp]
\vspace{-10pt}
\caption{Ablations on KARMA. We replaced or removed different decomposition components. The results reported here are the \textbf{averaged} performance across all forecasting window $T \in \{96, 192, 336, 720\}$, given a fixed look-back window $L = 96$.}
\label{tab:ablation}
\centering
\begin{small}
    \renewcommand{\multirowsetup}{\centering}
    \setlength{\tabcolsep}{6pt}
    \renewcommand{\arraystretch}{0.5}
    \resizebox{0.8\columnwidth}{!}{
    \begin{tabular}{c|c|c|cc|cc|cc}
    \toprule
     \multirow{2}{*}{Component} &\multirow{2}{*}{$ATCD$} & \multirow{2}{*}{$HFTD$} & \multicolumn{2}{c|}{Weather} & \multicolumn{2}{c|}{ECL} & \multicolumn{2}{c}{ETTm1}\\
    \cmidrule(lr){4-5} \cmidrule(lr){6-7}  \cmidrule(lr){8-9}
    & & & MSE & MAE & MSE & MAE  & MSE & MAE\\
    \midrule
    \multirow{4}{*}{\textbf{KARMA$_{w/o}$}} & \Checkmark & \Checkmark & \textbf{0.250} & \textbf{0.277} & 0.168 & \textbf{0.261} & \textbf{0.387} & \textbf{0.399}\\
    & \XSolidBrush & \Checkmark & 0.257 & 0.279 & \textbf{0.163}& 0.262 & 0.391 & 0.404\\
    & \Checkmark & \XSolidBrush & \textbf{0.250} & 0.280 & 0.193 & 0.295 & 0.395 & 0.405\\
    & \XSolidBrush & \XSolidBrush & 0.263 & 0.285 & 0.179 & 0.269 & 0.396 & 0.407\\
    \midrule
    \multirow{2}{*}{\textbf{KARMA$_{replace}$}} & $STL$ &   & 0.258 & 0.281 & 0.174 & 0.265 & 0.389 & 0.403\\
     &  & $TFBlock$ & 0.255 & 0.283 & 0.183 & 0.276 & 0.393 & 0.404 \\
    \bottomrule
    \end{tabular}
    }
\end{small}
\vspace{-10pt}
\end{table}
An interesting observation is that on the ECL dataset, the model exhibits a significant dependency on HFTD, whereas the reliance on ATCD is minimal. We speculate that this is because the ECL dataset has a highly periodic trend, making the focused processing of seasonal components less critical.

Additionally, the replacement experiments reveal that STL decomposition shows significant shortcomings, whereas the KarmaBlock, based on Mamba, significantly outperforms the Transformer-based TFBlock in time series analysis.


\subsubsection{Efficiency.}
\begin{figure}[!h]
\vspace{-30pt}
\centering
\includegraphics[width=0.8\columnwidth]{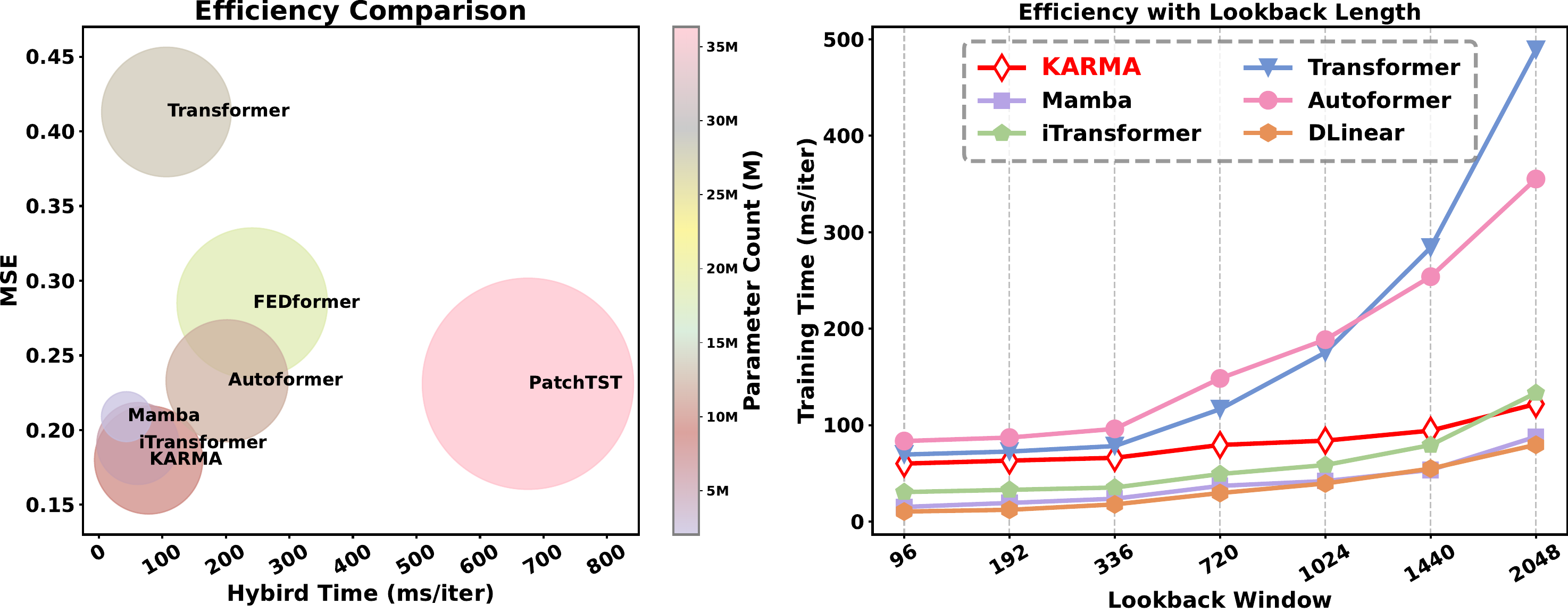}
\caption{Model efficiency comparison under input-720-predict-720 of ECL (left) . Comparison of training efficiency corresponding to different input and output lengths (right)}
\label{fig5_efficiency}
\vspace{-10pt}
\end{figure}

We analyzed the efficiency of KARMA on the ECL dataset from two perspectives. Fig. \ref{fig5_efficiency} (left) compares the model parameter count, training time, and MSE accuracy under the condition of $L = T = 720$. Compared to Transformer-based methods, KARMA achieves efficient computation and precise forecasting results while maintaining a smaller parameters. 

Fig. \ref{fig5_efficiency} (right) illustrates training speeds across various input-output lengths ($L = T \in \{96, 192, 336, 720, 1024, 1440, 2048\}$). It can be observed that the training time of Transformer-based methods grows quadratically with sequence length (notably, iTransformer achieves near-linear growth through special inversion processing but shows performance degradation for ultra-long sequences with $L \geq 2048$). In contrast, KARMA demonstrates a linearly and gradually increasing training time as sequence length grows. While DLinear exhibits exceptional efficiency due to its simple MLP architecture, there is still significant room for improvement in its prediction accuracy. 

\section{Conclusion}
The paper proposes the KARMA framework, which leverages ATCD and HFTD multilevel decomposition techniques, combined with the multi-scale Mambas stacked together to form KarmaBlocks to achieve efficient multivariate long-term time series forecasting. ATCD dynamically extracts trend and seasonal components, while HFTD employs discrete wavelet transform to decompose frequency and temporal information. These are then processed through multiple KarmaBlocks for multi-scale modeling, enabling comprehensive feature capture. Experimental results demonstrate that KARMA outperforms existing methods across various real-world datasets, particularly on data with pronounced trend and seasonal characteristics and achieves state-of-the-art performance. In addition, we also verified its adaptability and efficiency. In the future, we will conduct research on large-scale time series models for few-shot and general purposes.

\begin{credits}
\subsubsection{\ackname} This work was supported by the Special Project on Square Kilometre Array (SKA) of the Ministry of Science and Technology of China (Grant No. 2020SKA0120202) and the National Natural Science Foundation of China (Grant No. 12373113 and No. 12475196).

\end{credits}

%
%
%
\bibliographystyle{splncs04}
\bibliography{reference}
%
\end{document}